\documentclass[pdflatex,sn-mathphys-num]{sn-jnl}

\usepackage{graphicx}%
\usepackage{multirow}%
\usepackage{amsmath,amssymb,amsfonts}%
\usepackage{amsthm}%
\usepackage{mathrsfs}%
\usepackage[title]{appendix}%
\usepackage{xcolor}%
\usepackage{textcomp}%
\usepackage{manyfoot}%
\usepackage{booktabs}%
\usepackage{algorithm}%
\usepackage{algorithmicx}%
\usepackage{algpseudocode}%
\usepackage{listings}%

\theoremstyle{thmstyleone}%
\theoremstyle{thmstyletwo}%

\theoremstyle{thmstylethree}%

\begin{document}

\title[Article Title]{scAgent: Universal Single-Cell Annotation via a LLM Agent}


\author[1]{\fnm{Yuren} \sur{Mao}}\email{yuren.mao@zju.edu.cn}

\author[1]{\fnm{Yu} \sur{Mi}}\email{miyu@zju.edu.cn}

\author[1]{\fnm{Peigen} \sur{Liu}}\email{peigenliu@zju.edu.cn}

\author[1]{\fnm{Mengfei} \sur{Zhang}}\email{zmengfei@zju.edu.cn}

\author[2]{\fnm{Hanqing} \sur{Liu}}\email{
hanqingliu@g.harvard.edu}

\author*[1]{\fnm{Yunjun} \sur{Gao}}\email{gaoyj@zju.edu.cn}

\affil*[1]{\orgdiv{School of Software Technology}, \orgname{Zhejiang University}, \orgaddress{\city{Hanghou}, \postcode{310058}, \state{Zhejiang}, \country{China}}}

\affil[2]{\orgdiv{Society of Fellows}, \orgname{Harvard University}, \orgaddress{\city{Cambridge}, \postcode{02138}, \state{Massachusetts}, \country{USA}}}








\abstract{Cell type annotation is critical for understanding cellular heterogeneity. Based on single-cell RNA-seq data and deep learning models, good progress has been made in annotating a fixed number of cell types within a specific tissue.
However, 
universal cell annotation, which can generalize across tissues, discover novel cell types, and extend to novel cell types, remains less explored. To fill this gap, this paper proposes scAgent, a universal cell annotation framework based on Large Language Models (LLMs). scAgent can identify cell types and discover novel cell types in diverse tissues; furthermore, it is data efficient to learn novel cell types. Experimental studies in 160 cell types and 35 tissues demonstrate the superior performance of scAgent in general cell-type annotation, novel cell discovery, and extensibility to novel cell type.
}




\maketitle

\section{Introduction}\label{sec1}

Cell type annotation (CTA), which aims to identify the type of given cells, is a fundamental step of single-cell RNA sequence (scRNA-seq) analysis. It can resolve cellular heterogeneity across cell populations and help us better understand the gene functions in health and disease \cite{cheng2023review}. Given a set of unannotated cells, we can annotate their cell types based on manually curated marker genes or annotated reference data. The marker genes, which typically have specific expression on corresponding cell types, can directly indicate the cell types. However, there are some cells of a cell type that do not have a high expression in their corresponding marker genes \cite{pullin2024comparison}, and the marker genes are incomplete to identify all cell types \cite{chen2024sccts}. Furthermore, selecting marker genes for cell types is manual and time consuming. Thus, the effectiveness and efficiency of marker gene-based methods cannot meet the need for large-scale cell-type annotation tasks, such as tasks in the Human Cell Atlas (HCA) \cite{human_cell_atlas}.
By contrast, the annotated reference data are rapidly accumulating, which covers more and more tissues and cell types. It can be used as training data for learning CTA classifiers that annotate cell types without human intervention. Based on deep neural networks, several CTA classifiers have been proposed and performed high effectiveness and efficiency. 

However, most of the existing CTA models \cite{celltypiest,geneformer, tosica, scbert, scgpt, scfoundation, uce, scautomic, scinterpreter, schyena} are not generalize well across diverse tissues. To address this issue,  recently scTab \cite{sctab} is proposed to learn a cross-tissue classifier on large data sets. Nevertheless, it is  cannot discover and extend to novel cell types. Moreover, it is data-inefficient. The universal cell annotation, which can generalize across tissues, discover and extend to novel cell types, remains less explored. To achieve universal cell annotation, this paper propose a cell annotation agent framework based on Large Language Models (LLMs), dubbed scAgnet. It consists of a planning module, a memory module, and a tool hub. Given scRNA-seq and extra information discribed by natural language, scAgnet automatically annotates the scRNA-seq after a multi-turn interaction between the three modules. Furthermore, it can automatically detect the unknown cell types, easily extending to novel types.  Extensive experiments on CELLxGENE\cite{cellxgene} and Tabula Spaiens\cite{tabula_sapiens} illustrate that it has superior CTA performance compared with existing methods.

\section{Results}\label{sec2}

\begin{figure}[H]
\centering
\includegraphics[width=\textwidth]{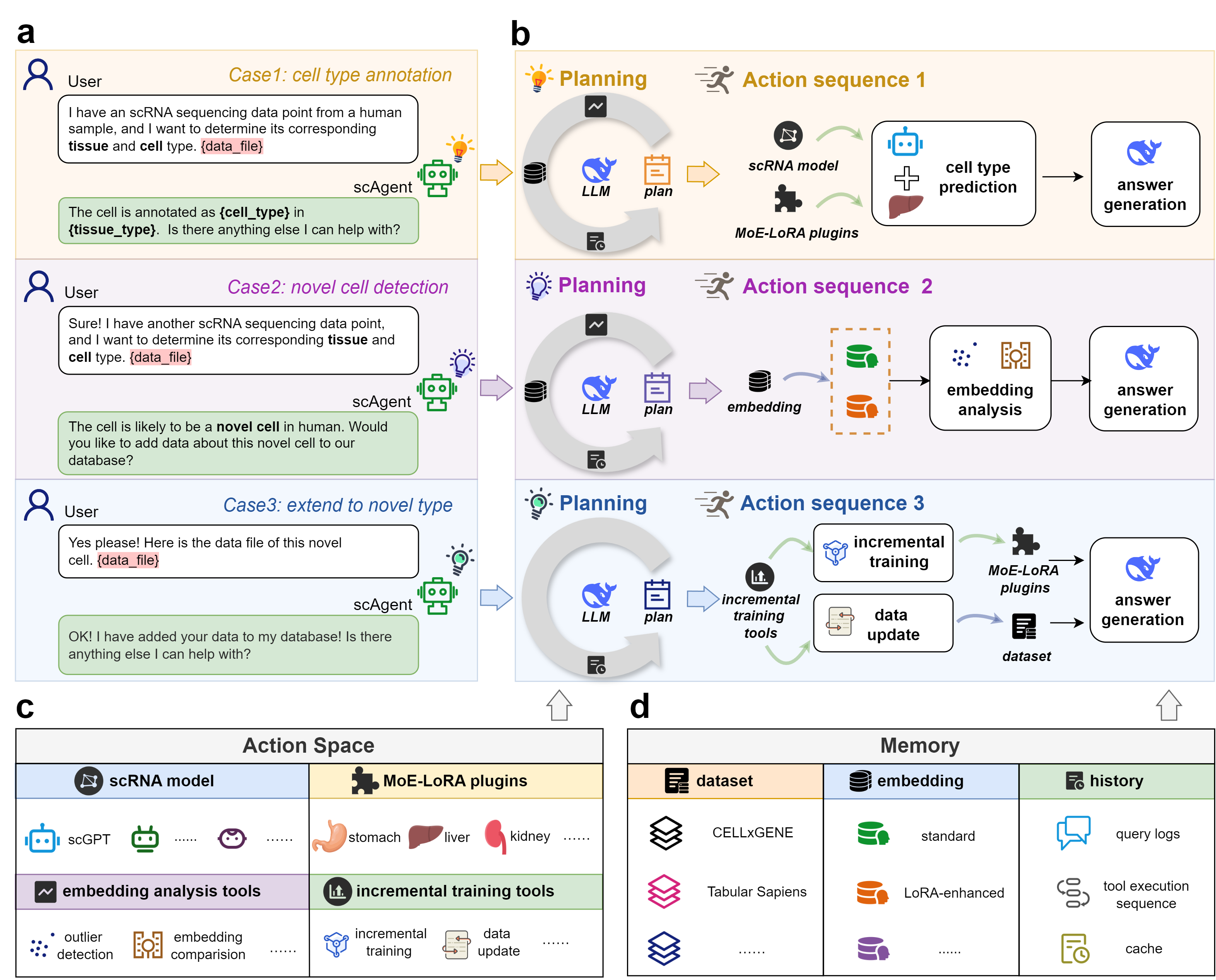}
\caption{\textbf{Overview of scAgent.} \textbf{a} Simulation of various user queries. scAgent can generate proper answers according to different user requests, including cell type annotation, novel cell detection and extension to novel type. \textbf{b} The planning module of scAgent. The planning module receives user query and generates a plan as output. The planning process is primarily driven by LLMs(DeepSeek-R1 671B), and assisted by the tools from the action space and the information from memory module, which are illustrated as black icons on the circular arrow. The generated plan determines the action sequence. In the action sequence, the black icons represent a certain category of tools or memory, which can be found in the action space and the memory module. The green arrows denote the interaction with the action space, while the dark blue arrows refers to the memory module. The white box signifies an action, achieved through the collaboration of one or multiple tools, and integrated with memory as needed. \textbf{c} The composition of action space. scAgent employs scGPT(pre-trained on 33 million cells) as the foundational scRNA model while maintaining extensibility to other deep learning models. There are over 30 MoE-LoRA plugins of specific tissues. The embedding analysis tools consists of outlier detection and embedding comparison. Through the analysis of outliers and by comparing the input embeddings with the embeddings stored in memory, these tools can assist with cell annotation tasks and the discovery of novel cell types. The incremental training tools includes training and data update tools. The data update tool merges the original data and new data, also support the update of datasets and other corresponding information in the memory module. The incremental training tool can continually train the MoE-LoRA plugins. \textbf{d} The information in the memory module. The published datasets are stored in the memory module for model training. Embeddings are categorized as LoRA-enhanced and standard, which refer to hidden states that generated by scRNA models with or without MoE-LoRA plugins respectively. System history includes query logs, tool execution sequence and cache, which can help with efficiently planning.}\label{scagent_overview}
\end{figure}

\subsection{The Framework of scAgent}

scAgent is an LLM-based autonomous agent\cite{agent-survey} designed for universal cell annotation. The system comprises three core components: a planning module, an action space, and a memory module. Users submit queries and data files through a natural language interface. Upon receiving inputs, scAgent leverages the planning module to formulate an execution plan by integrating tools from the action space and knowledge from the memory module. Based on this plan, scAgent constructs and executes an action sequence to derive the final result. The system ultimately returns a structured natural language answer synthesized by the LLM. 
As demonstrated in Fig. \ref{scagent_overview}a, scAgent adapts effectively to diverse annotation scenario. This universal capability relies on the integration of LLM-powered planning module and an extensible action space. Facilitated by the intelligence of LLM, scAgent is able to discover novel cell, meanwhile enhancing cell annotation performance. By integrating the modular and scalable tools in the action space, scAgent achieves cross-tissue generalization as well as extension to novel cell types.

The planning module is the core of scAgent. As illustrated in Fig. \ref{scagent_overview}b, this module accepts user queries and scRNA-seq data files as input, and generates detailed plans to guide the execution of action sequences. Its planning ability derives from the inference capability of LLMs driven by predefined prompts. Specifically, we employ DeepSeek-R1 671B as the planning LLM due to its strong inference ability, ensuring high-quality plan generation. Additionally, the planning module utilizes tools from the action space and information from the memory module to enhance the planning process. When necessary, it refines the plan through multiple iterations. This sophisticated architecture allows scAgent to handle complex scenarios with robust and context-aware decision-making.

To execute the generated plan, scAgent is equipped with an action space (Fig. \ref{scagent_overview}c). The action space contains a diverse set of tools, including scRNA models, MoE-LoRA plugins, incremental training tools, and embedding analysis tools. 
scRNA models are deep learning models pretrained on scRNA-seq datasets. Specifically, we use scGPT as the primary scRNA model, which has been trained on over 33 million cells. scGPT is a pretrained transformer encoder over 33 million cells, which is used as an embedding model to encode a cell's gene expression profile into a latent representation for cell similarity assessment. For cell type annotation tasks, we augment the base scGPT model with a multilayer perceptron (MLP) classification head.
To enhance scGPT's ability on cross-tissue annotation, we further implement MoE-LoRA architecture to fine-tune scGPT.
MoE-LoRA integrates Low-Rank Adaptation (LoRA)\cite{lora} with Mixture-of-Experts\cite{moe}, where each LoRA module serves as a specialized expert. MoE-LoRA plugin is a pluggable parameter module that dynamically integrates with the pre-trained scRNA model’s weights. By sharing the base model’s pretrained weights and fine-tuning only low-rank adapter layers, the architecture archieves high data efficiency, with the plugins only require a small number of tissue-specific training data. During annotation, scAgent automatically loads the corresponding plugin for a target tissue, enabling tissue-adapted predictions. When new data becomes available, MoE-LoRA architecture is able to support incremental plugin updates while preventing catastrophic forgetting through isolated plugin parameters. This design also ensures infinite scalability, allowing continuous addition of new plugins.
To support extensible updates, we integrate incremental training tools. These tools firstly mix new data with existing datasets, then retrain the specific tissue plugin using the combined data, finally update relevant information, including new data and plugins. 
Furthermore, embedding analysis tools assist in cell annotation and novel cell detection. Here, embedding refers to the intermediate output of classifier models, which represent the feature of input cell. By analyzing outlier or comparing input embedding with memory embeddings, these tools enhance the performance of cell annotation or novel cell detection. 
Together, the tools in the action space ensure scAgent’s high performance, adaptability, and efficiency across various tasks.

The memory module serves as a dynamic knowledge repository(Fig. \ref{scagent_overview}d). It stores datasets, embeddings, and system history. Datasets include published datasets such as datasets on CELLxGENE\cite{cellxgene} and Tabula Spaiens\cite{tabula_sapiens}, as well as continuously accumulating user-uploaded data. These datasets support MoE-LoRA plugin training.
Embeddings are the representations of cell features, which are used to compare the input data feature with the exising features, subsequently assisting with the novel cell detection process.  We generate embeddings using scRNA models with and without MoE-LoRA plugins, providing multiple feature perspectives for embedding analysis. These embeddings are stored in the Milvus vector database for efficient retrieval.
The memory module also records system history, including query logs, tool execution sequences, and cache data. These historical records assist in decision-making and enhance system efficiency. 
The structured design of the memory module enables efficient management of large-scale data while preserving essential contextual information for diverse annotation tasks.

\subsection{ScAgent Enables Data-efficient Cross-tissue CTA}\label{subsec2.2}

\begin{figure}[H]
\centering
\includegraphics[width=\textwidth]{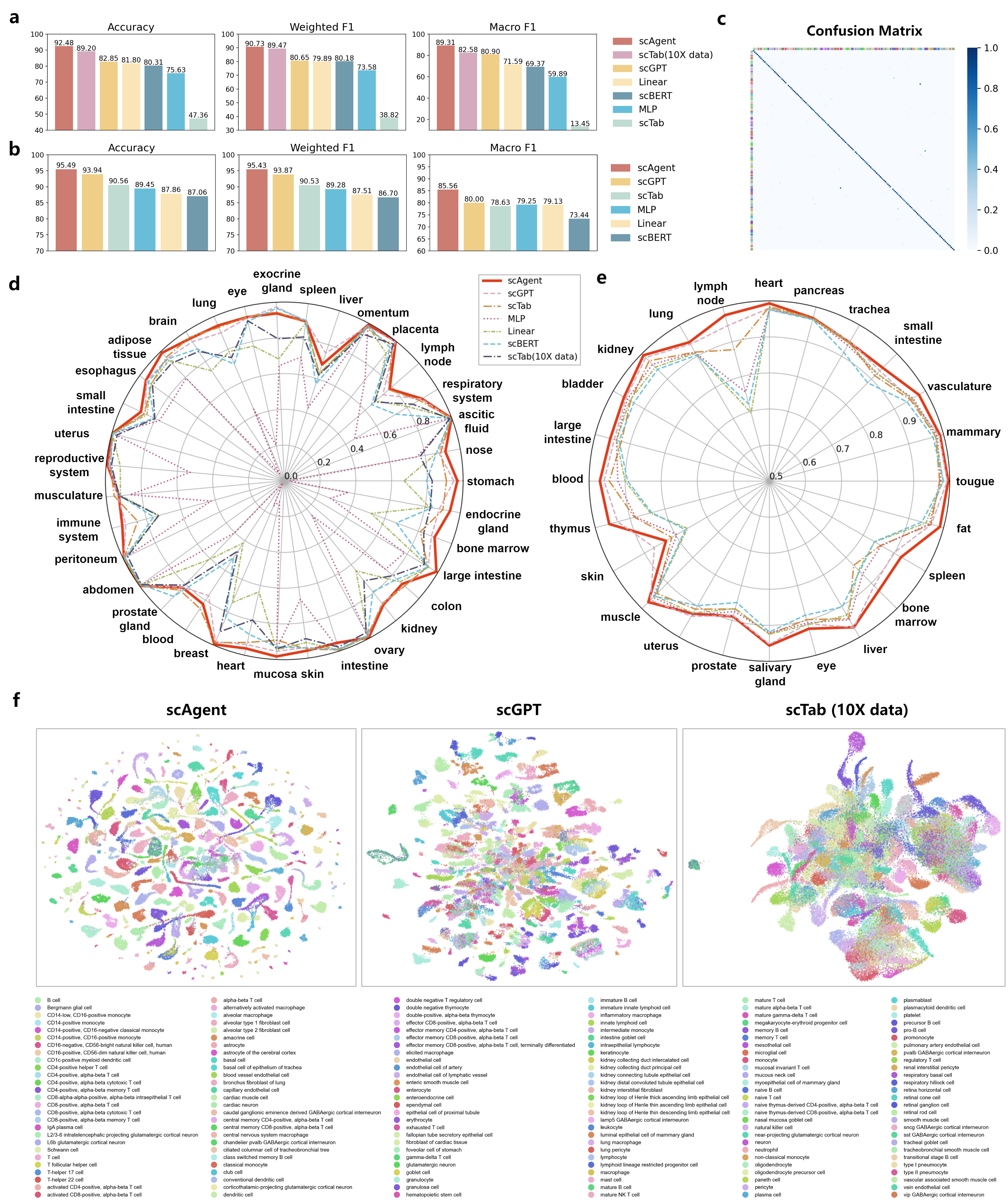}
\caption{\textbf{Cross-tissue CTA results of scAgent.} \textbf{a,b} scAgent ranks first on accuracy, weighted F1-score, and macro F1-score compared to other CTA methods on CG dataset (\textbf{a}) and TS dataset (\textbf{b}). The bars in the bar chart are arranged in order from highest to lowest. \textbf{c} scAgent can specify diverse cell types. In the confusion matrix, each row represents the true cell type in CG reference dataset, and each column represents the predicted cell type by scAgent. The color coding for the cell types is provided in the legend below. The values in the confusion matrix have been normalized by row, such that each value represents the recall rate for the corresponding true cell type. \textbf{d,e} scAgent shows superior tissue-specific performance on CG dataset (\textbf{d}) and TS dataset (\textbf{e}). Each vertex of the radar chart represents a specific tissue, and the length of the axis indicates the weighted F1-score for cell annotation performance on this tissue. \textbf{f} scAgent captures the distinctive features of diverse cell types. Compared to scGPT and scTab (10X data), the UMAP visualization of scAgent on CG reference dataset shows greater distances between cell clusters, demonstrating its superior feature extraction capability.}\label{sctab_ts3}
\end{figure}

We test scAgent across 35 human tissues and compare it with other six methods. scAgent demonstrates superior performance. The scRNA-seq dataset used in the comparison is randomly drawn from CELLxGENE\cite{cellxgene}, which contains 1 million cells spanning 35 human tissues and 162 cell types. In the following sections, we will abbreviate it as CG dataset.We split CG dataset into reference set (90\%) and query set (10\%). The other six methods are scGPT\cite{scgpt}, scTab (10X data)\cite{sctab}, scTab\cite{sctab}, scBERT\cite{scbert}, MLP and linear classifier respectively. In scGPT, we add an MLP layer as the classification head to enable it to perform the 162-class multi-classification task. scTab is trained following the exact configuration described in \cite{sctab}. scTab (10X data) employs the model weights from \cite{sctab}, which are trained on a dataset that is homologous to CG dataset but approximately 10 times larger in size. The classification head of scBERT is also extended to accommodate this CTA task. As for MLP, we set the number of hidden layers to 2 to prevent overfitting during training. Linear strictly adhere to the configuration specified in \cite{sctab}. 
To ensure fairness, these methods are all evaluated with three metrics: accuracy, macro F1-score, and weighted F1-score.

scAgent achieves state-of-the-art performance with the three metrics (Fig. \ref{sctab_ts3}a). Its macro F1-score is 89.31\%, surpassing the second-best method scTab (10X data) by 6.73 percentage points. However, the number of training data of scAgent is ten times less than that for the scTab (10X data). It demonstrates scAgent is more data-efficient. Besides, we illustrate the tissue-sepcific performance of scAgent using radar chart (Fig.\ref{sctab_ts3}d), from which we can see that scAgent shows a global superiority. Notably, scAgent achieves near-ceiling weighted F1-scores ($>$ 99\%) in 8 critical tissues, including the uterus, placenta, and breast. Additionally, we calculate the standard deviation of weighted F1-scores across all tissues for these methods. The standard deviations for scAgent, scGPT, and scTab (10X data) are around 0.07, while those for other baselines exceed 0.1. This indicates that scAgent not only maintains leading performance but also ensures consistency and robustness in different tissues.

Moreover, scAgent demonstrates exceptional performance in distinguishing diverse cell types. As shown in the confusion matrix (Fig. \ref{sctab_ts3}c; detailed values available in supplementary file; the diagonal values represents recall rates), scAgent achieves remarkable performance, with over 90\% recall for 84\% of the cell types. Specifically, it excels in identifying fine-grained cell subtypes, such as cortical interneurons, with recall rates of 96.3\% for pvalb and 95.1\% for sst subtypes. In addition, comparative UMAP visualization (Fig. \ref{sctab_ts3}f) against scGPT and scTab (10X data) substantiates scAgent's superior clustering capability across different cell types. The superior performance of scAgent stems from two key components: (1) The MoE-LoRA plugins fine-tuning on scRNA models enhances the model's ability to structure the latent space according to cell type labels, leading to more discriminative embeddings; (2) The agentic framework leverages LLM's decision-making capability to dynamically select and chain together the most appropriate MoE-LoRA modules for different tasks. 

Further evaluation on the Tabula Sapiens dataset\cite{tabula_sapiens} proves scAgent's cross-dataset generalization capability. The Tabula Sapiens dataset is a comprehensive human reference cell atlas comprising nearly 500,000 cells spanning 24 distinct tissues, which we will abbreviate as TS dataset in subsequent sections. Unlike CG dataset, TS dataset offers finer-grained annotations, providing deeper insights into cell states and differentiation processes across various tissues. Despite the inherent differences between these two datasets, scAgent consistently delivers outstanding performance across the three metrics on TS dadaset(Fig. \ref{sctab_ts3}b). Moreover, scAgent also shows superior tissue-sepcific performance in the radar chart (Fig. \ref{sctab_ts3}e). It achieves near-ceiling weighted F1-scores (>99\%) in 4 tissues including tougue, kidney, heart and mammary. The consistent results across different datasets underscore the strong generalization capability of scAgent.

\subsection{ScAgent Enables Discovery of Novel Rare Cell Types even Batch Effect Occurs}\label{subsec2.5}

To achieve universal cell annotation, it is necessary to be able to discover novel rare cell types that have not been seen in the reference data (e.g., cancer cells).  Discovering novel rare cell types  is important for biology research and precision medicine. 
To evaluate scAgent can effectively identify the novel cell types, we adopt two dataset: 
   (i) a dataset sampled from liver tissue of breast cancer patients (hereafter referred to as Liver Breast Cancer) \cite{KRISHNA2021662}, and (ii) a dataset sampled from kidney tissue of clear cell renal cell carcinoma (ccRCC) patients (hereafter referred to as Kidney ccRCC) \cite{klughammer2024multi}.
For the Liver Breast Cancer dataset, 
the known cell types that have been contained in the reference data are: T cells, endothelial cells, macrophages, and monocytes, while the novel cell type that has not been contained by the reference data is malignant cell.
Besides, for the Kidney ccRCC dataset, The known cell type is endothelial cell, while the novel cell type is the abnormal cell.

\begin{figure}[H]
\centering
\includegraphics[width=1\textwidth]{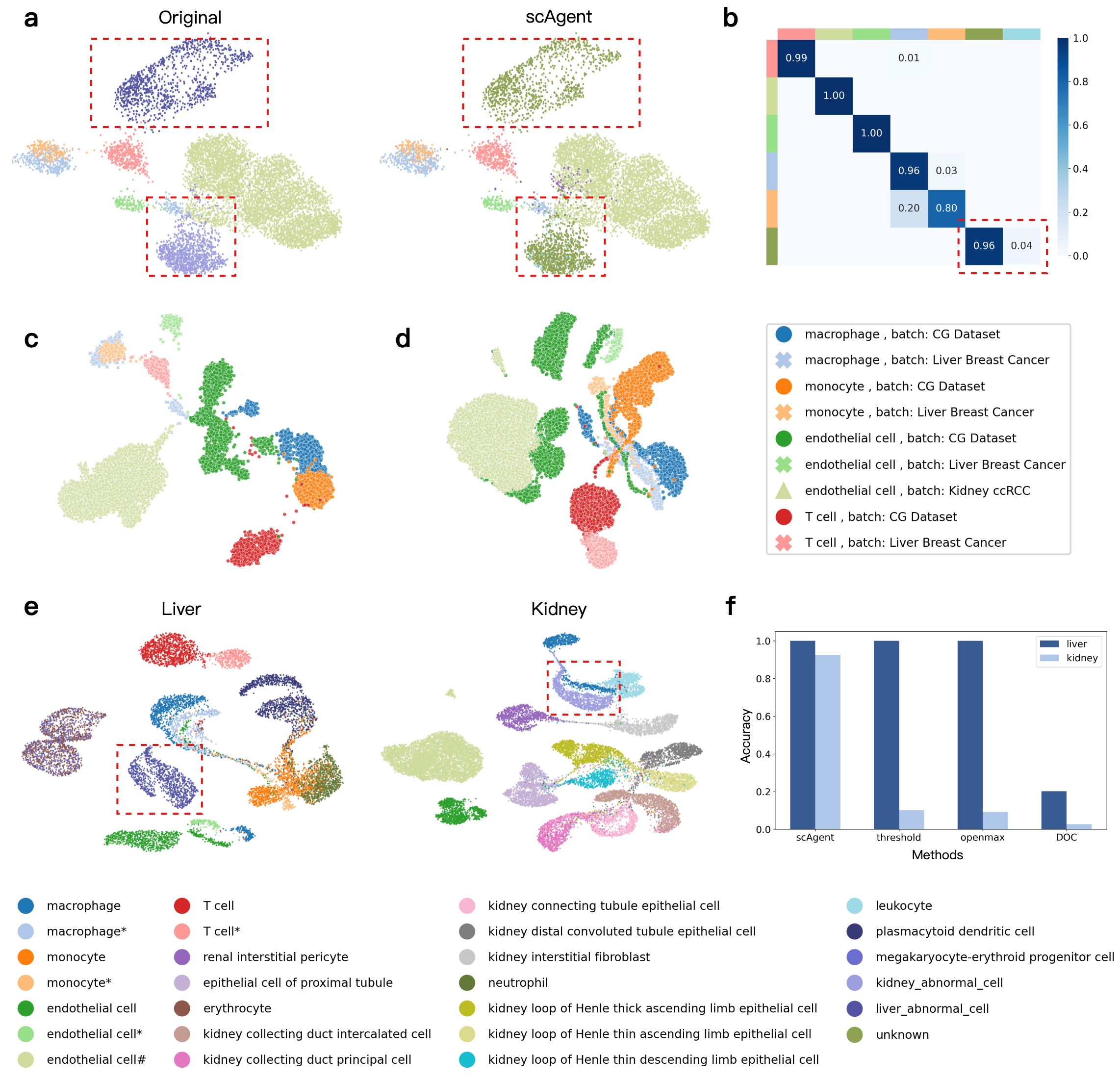}
\caption{\textbf{Performance of scAgent in novel cell detection and batch effect correction.} \textbf{a} UMAP visualization of the raw feature space for the Liver Breast Cancer and Kidney ccRCC datasets. Novel cells are clustered separately from normal cells and are highlighted by red bounding boxes.
 \textbf{b} scAgent labels most of novel cells  as unknown. Heatmap shows proportion of cells in each row with original label O (shown on the right) predicted as cell type P (shown on the top).
 \textbf{c} UMAP visualization of the raw feature space for data from different batches. The same cell type from different batches is widely separated, indicating significant batch effects.
\textbf{d} UMAP visualization of features of different dataset batches, provided by scAgent. The same cell type from different batches is clustered, demonstrating the effectiveness of scAgent in reducing batch effects. 
 \textbf{e} UMAP visualization of features of Liver Breast Cancer (left panel, marked by *) and Kidney ccRCC (right panel, marked by \#) compared to CG reference data, provided by scAgent. Novel cells (red circled) in the liver are well-separated from reference data, while those in the kidney overlap with leukocytes and macrophages, making detection more challenging.
 \textbf{f} Novel cell detection accuracy of scAgent and three other methods (threshold, OpenMax, DOC) on  Liver Breast Cancer and Kidney ccRCC datasets. scAgent outperforms other methods consistently.}
 \label{openset}
\end{figure}

In the raw feature space of  these two datasets, the novel cells are clustered separately from the known cells, shown as the corresponding UMAP where unknown cells are marked by the red bounding boxes (Fig. \ref{openset}a). scAgent can annotate  96\% of novel cells as unknown (Fig. \ref{openset}b). Furthermore, scAgent can accurately annotate the cells belonging to the known types, the accuracy are 99\%, 100\%, 96\% and 80\% respectively (Fig. \ref{openset}b). It demonstrates scAgent is robust to batch effects  for these two datasets are generated on  experimental platforms that are different from the reference data. The UMAP for the raw feature space visually illustrates the batch effects (Fig. \ref{openset}c), while the feature space for scAgent demonstrates scAgent can successfully annotate cell types with the existence of batch effects (Fig. \ref{openset}d). 

Besides, scAgent performs consistently over different cell type discovery tasks that have different degree of difficulty. To showcase this, we compare scAgent with three widely used novel cell discovery methods on the Kidney ccRCC and Liver Breast Cancer respetively. The  three methods are (1) threshold, which rely on predefined cutoffs for identifying unknown cells; (2) OpenMax, which extends the softmax layer of a neural network to estimate the probability of an input belonging to an unknown class\cite{openmax}; (3) DOC (Deep Open Classification), which replaces the softmax output layer with a 1-vs-rest sigmoid layer to enhance out-of-distribution detection\cite{DOC}. The unknown cells in the Liver Breast Cancer dataset are clustered separately from the cells in the reference data, while the cluster of unknown cells in the Kidney ccRCC dataset overlaps with clusters of leukocytes and macrophages in the reference data (Fig. \ref{openset}e). Detecting unknown cells in the Kidney ccRCC dataset is more difficult than that for the Liver Breast Cancer dataset. Both scAgent and the other two methods achieve good performance on the Liver Breast Cancer dataset. However, the other three methods perform poorly on the Kidney ccRCC dataset. The accuracy for scAgent is 92.6\%, while threshold, OpenMax, and DOC only achieve 10.1\%, 9.2\%, and 2.7\% (Fig. \ref{openset}f).

\subsection{ScAgent Enables Efficient Incremental Learning for Novel Cell Types}\label{subsec2.6}

After discovering a novel cell type, a universal cell type annotation framework should be able to learn this novel type and annotate cells belonging to this type next time. It requires the cell type annotation framework supports incremental learning, which is ignored by existing cell type annotation methods \cite{celltypiest,geneformer, tosica, scbert, scgpt, scfoundation, uce, scautomic, scinterpreter, schyena}. They cannot extend to novel cell types without changing their network architecture. We demonstrate that scAgent supports effectively incremental learning and can be easily extened to novel cell types. Specifically, we showcase scAgent can effectively extend to the malignant tumor cells discovered in the last section.

\begin{figure}[H]
\centering
\includegraphics[width=\textwidth]{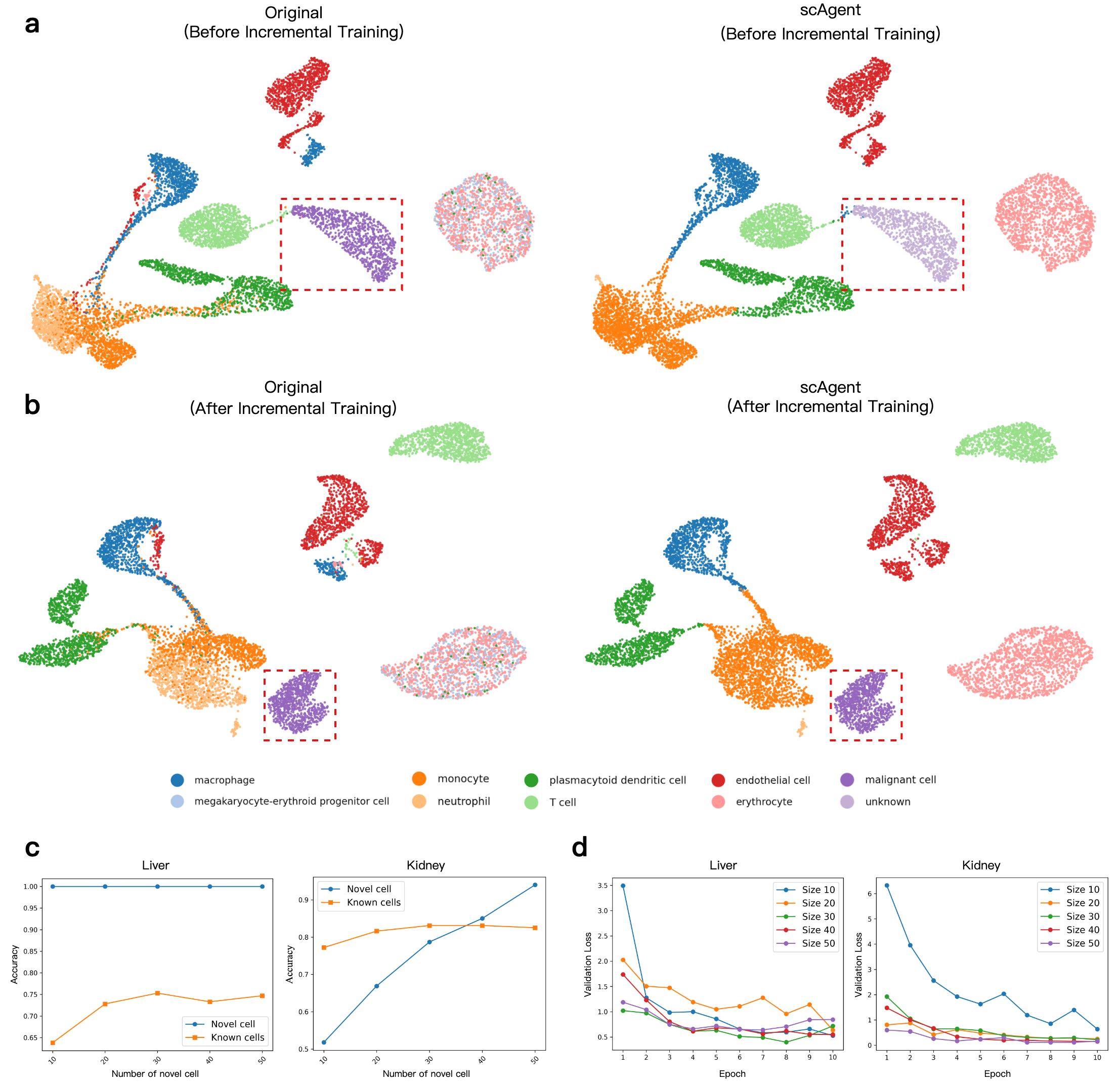}
\caption{\textbf{Results of incremental training in scAgent.} \textbf{a} The UMAP visualization is generated based on the embeddings of reference data from liver tissue and malignant tumor cells prior to incremental training. The malignant tumor cells, highlighted with red bounding boxes, are classified as "unknown" by scAgent since they were not present in the reference data.
\textbf{b} Following incremental learning with a limited number of labeled samples, scAgent successfully annotates these previously unknown cells with their ground truth labels (indicated by red bounding boxes).
\textbf{c} We vary the number of incremental learning samples (i.e., unknown cells with ground truth labels) from 10 to 50 and evaluate the recognition accuracy for both novel cells and known cells.
\textbf{d} The cross-entropy loss on the validation set is plotted after each training epoch for scAgent, demonstrating the model's performance after incremental learning with a few number of labeled data.}\label{increment}
\end{figure}

As demonstrated in the preceding section, scAgent is capable of annotating malignant tumor cells in liver and abnormal cells in kidney as unknown for they has not appeared in the reference data(Fig. \ref{increment}a). Here, we show that scAgent can correctly annotate these unknown cells as their groud truth labels after incremental learning with a few number of labeled data (Fig. \ref{increment}b). To 
verify the incremental learning for scAgent is data-efficient, we vary the number of incremental learning data (i.e., unknown cells with ground truth labels) from 10 to 50 and report the corresponding performance of scAgent's incremental learning (Fig. \ref{increment}c).
For the liver breast cancer data set, scAgent can achieve accuracy 100\% for malignant cells when the number of labeled data is equal to or greater than 30. In addition, the incremental learning has not had a negative influence on the performance of known cell types when the number is equal to or greater than 30. Before the incremental learning, the annotation accuracy on the known cell types  74.9\% for liver, which is comparable with the accuracy obtained after incremental learning. For the Kidney ccRCC dataset  where the unkown cells is more difficult to detect, scAgent can achieve accuracy over 80\%  for both the abnormal and known cell types when the number of incremental learning data is equal to or greater than 30. Besides, during incremental training, scAgent converges quickly  when the number of incremental learning data is equal to or greater than 30 (Fig. \ref{increment}d). These results verify that scAgent enables efficient incremental learning for novel cell types.

\section{Discussion}\label{sec12}
We introduce scAgent, a universal LLM-based autonomous agent for universal cell annotation. By integrating an intelligent planning module, an extensible action space, and a dynamic memory module, scAgent achieves cross-tissue generalization, novel cell type discovery, and efficient incremental learning. It outperforms existing methods in accuracy, macro F1-score, and weighted F1-score across diverse tissues and datasets, even under batch effects. Leveraging existing tools in the action space, scAgent demonstrates robust and scalable performance, making it powerful for precise analysis of scRNA-seq data. For future work, we plan to expand scAgent's capabilities in two key directions: (1) Incorporating multi-omic, spatial and perturbation data to enhance cell-type annotation through additional LoRA plugins that learn cross-modal relationships; and (2) Extending to diverse downstream tasks such as differential expression analysis, trajectory inference, and cell-cell interaction prediction. We will further explore in-context learning for zero-shot adaptation. In summary, scAgent represents a significant advancement in universal cell annotation, with great potential for leveraging single-cell data in biological discovery.

\section{Methods}\label{sec3}

\subsection{The Framework of scAgent}\label{subsec2.1}

scAgent is an advanced autonomous agent system built upon LLMs, specifically designed for universal cell annotation, including cross-tissue cell annotation, novel cell detection and extension to novel cell types. The system architecture comprises three core components: a planning module, an action space, and a memory module, which can be formally expressed as:

\begin{equation}
    \text{scAgent} = (P, A, M)
\end{equation}

where $P$ denotes the planning module, $A$ represents the action space, and $M$ signifies the memory module.

\subsubsection{The Planning Module of scAgent}

The planning module of scAgent is implemented based on the LangGraph framework \cite{langgraph}, which enables graph-based construction of agent systems. Its operational dynamics can be formalized as:

\begin{equation}
S_{t+1} = f(S_{t}, A_{t}, E_{t})
\end{equation}

where \( S_t \) represents the current system state, \( A_t \) denotes the action taken at time \( t \), and \( E_t \) encapsulates environmental information at time \( t \), which refers to the user query and the input scRNA-seq data. The state transition function \(f\) governs system evolution, with \(S\) encompassing tool hub and Memory states, and \(A\) including operations such as request analysis, tool invocation, information retrieval, and user interaction.

To enhance the inference capability of the planning module, we utilize DeepSeek-R1 671B\cite{deepseek} as the base LLM. 

\begin{figure}[H]
\centering
\includegraphics[width=\textwidth]{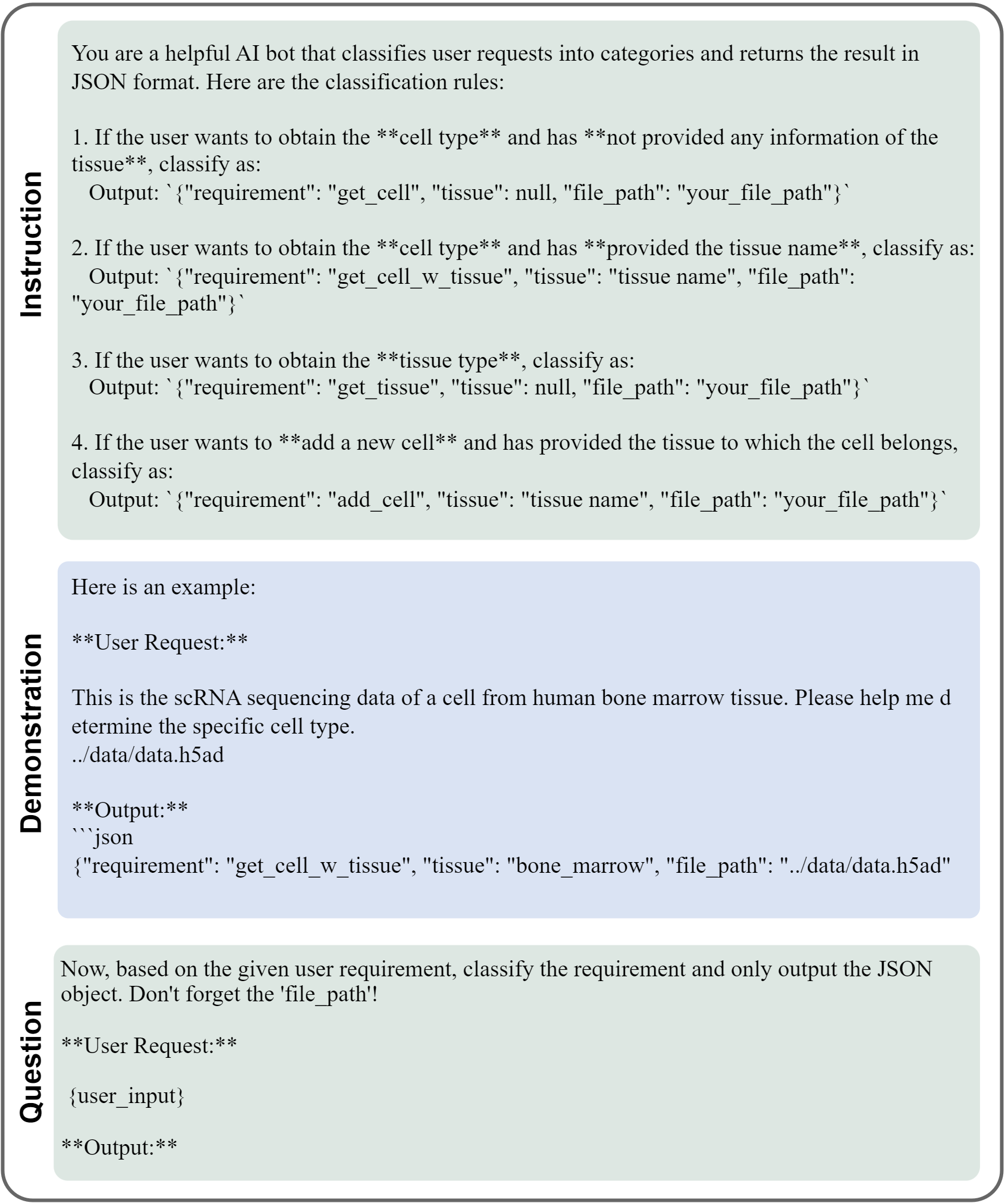}
\caption{Prompt Template for planning.}\label{prompt_input}
\end{figure}

\begin{figure}[H]
\centering
\includegraphics[width=\textwidth]{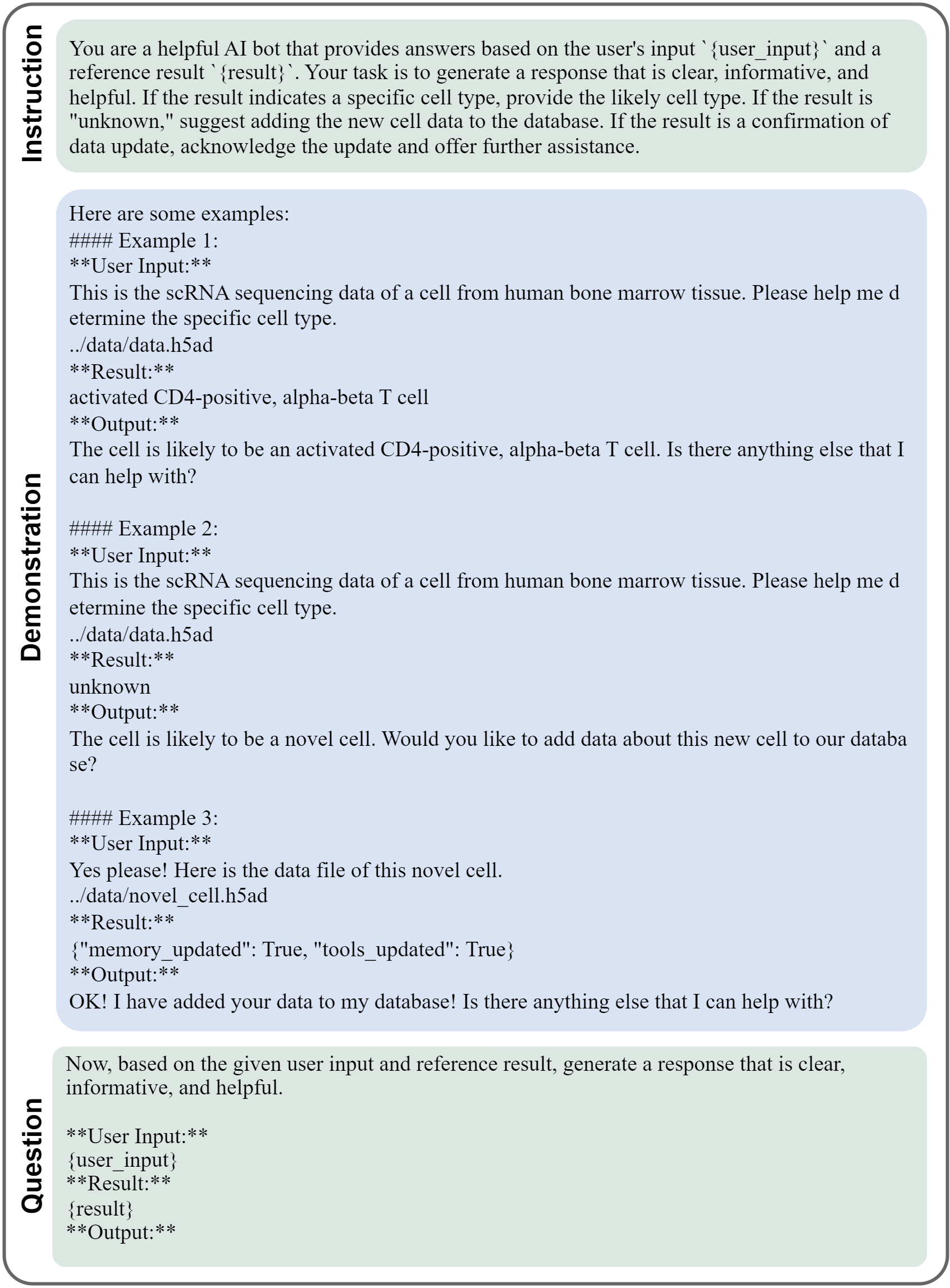}
\caption{Prompt Template for answer generation.}\label{prompt_output}
\end{figure}

Fig. \ref{scagent_overview}b illustrates three cases of planning. When scAgent receives a user query, it first leverages LLM to analyze user requirement, decides whether it's a cell annotation scenario or an extension to novel type scenario (Fig. \ref{prompt_input}). In the scenario of cell annotation, scAgent will process the input scRNA-seq data, using the scRNA model and the trained MoE-LoRA plugins from the action space to get input embeddings. After obtaining the embeddings, scAgent uses the embedding analysis tools to assess whether the cell could be a novel type. If the distance in the feature space is far enough, scAgent will regard the input cell as an existing type, and use the scRNA model and the trained MoE-LoRA plugins to predict the certain type (case 1). If the distance is close, scAgent will further use the LLM to compare the input embedding with the existing embeddings in the memory module, and then give a final decision on whether it's a nocel cell type or not (case 2). If the result is a novel cell type, scAgent will ask the user whether they would like to share their data with the memory module. If the user agrees, incremental training will proceed. In the scenario of extension to nocel type (case 3), scAgent updates dataset and embedding vectors in the memory module, as well as the MoE-LoRA plugins in the action space. Once the system detects the corresponding state updates, it will return it as a final result. Finally, scAgent use the LLM to organize and present the results in natural language to the user (Fig. \ref{prompt_output}).

\subsubsection{The Action Space of scAgent}

The action space constitutes the functional repository of scAgent, housing essential analytical tools categorized into four primary domains: scRNA models, MoE-LoRA plugins, embedding analysis tools, and incremental training tools (Fig.\ref{scagent_overview}c).

\paragraph{ScRNA Models}

scRNA models provide specialized CTA capabilities, predominantly employing Transformer-based deep learning architectures pre-trained on extensive scRNA-seq datasets. These models excel in capturing intricate gene expression patterns and generating robust feature representations for downstream analyses. scAgent utilizes scGPT\cite{scgpt} as the primary scRNA model, which implements generative pretraining techniques for precise cell type annotation and potential cell state characterization, thereby facilitating exploration of novel cellular phenotypes.

\paragraph{MoE-LoRA Plugins}

The MoE-LoRA plugins represent a sophisticated integration of Low-Rank Adaptation (LoRA) techniques \cite{lora} and Mixture of Experts (MoE) architecture \cite{moe}, forming the cornerstone of scAgent's dynamic adaptability.

LoRA introduces trainable low-rank decomposition matrices to efficiently fine-tune scRNA models. It updates model parameters as follows:

\begin{equation}
W = W_0 + \Delta W = W_0 + BA
\end{equation}

where \( W_0 \in \mathbb{R}^{d \times k} \) represents frozen original parameters, with \(d\) and \(k\) denoting input and output dimensions respectively. \( \Delta W \) represents the trainable parameters, which can be expressed as the product of two low-rank matrices \( B \in \mathbb{R}^{d \times r} \) and \( A \in \mathbb{R}^{r \times k} \), where \( r \) is the rank of the low-rank matrices. LoRA reduces the parameter update complexity from \( O(dk) \) to \( O(dr + rk) \), which is particularly effective when \( r \ll \min(d,k) \).

MoE architecture, originally proposed in \cite{moe0}, avoids interference between network layers in multi-task learning. In 2017, \cite{moe} introduced sparse gating to further optimize the computation process. MoE enables fast inference while maintaining large-scale models without significantly increasing computational cost. Specifically, MoE applies a gating network that dynamically activates specific "experts" (sub-modules) for each input, significantly improving model performance without proportional increases in computation. The formula is as follows:

\begin{equation}
y = \sum^{n}_{i=1}{g(x)_i \cdot e_i(x)}
\end{equation}

where \( n \) represents the number of "experts", \( g(x)_i \) is the gating network output for the \( i \)-th expert, and \( e_i(x) \) is the output of the \( i \)-th expert for input \( x \). A simple gating network \( g(x) \) can be represented as:

\begin{equation}
g(x) = \text{Softmax}(x \cdot W_r)
\end{equation}

In recent years, MoE-LoRA architecture is widely used in various works. As described in \cite{lora-survey}, this architecture is proper for cross-task generalization, thus can perform well in the cross-tissue CTA task. In MoE-LoRA architecture, each LoRA module acts as an expert. The gating network assigns mixed weights to each expert, allowing for flexible, plug-and-play use during prediction. The model parameter is updated as follows:

\begin{equation}
W = W_0 + \Delta W_{MoE-LoRA} = W_0 + \sum_{i=1}^{n}{g(x)_i \cdot B_i A_i}
\end{equation}

scAgent uses the MoE-LoRA architecture to implement an intelligent and efficient cell annotation system. It includes plugins for both tissue and cell type classification, catering to various user needs. Currently, the tool hub offers over 30 plugins, supporting multi-level biological annotation and continuous optimization of novel cell detection through embedding vector similarity retrieval. The pluggable design for parameter isolation ensures that adding new tissue types requires only extending independent plugins, avoiding the computational burden of full model retraining and mitigating catastrophic forgetting in incremental learning. This design, combining scalability and compatibility, makes the system suitable for large-scale single-cell RNA data analysis, providing cost-effective solutions for high-throughput analysis of millions of single cells.

\paragraph{Embedding Analysis Tools}

The embedding analysis tools in scAgent are essential for evaluating and interpreting the feature representations generated by the scRNA models. These tools focus on two key functionalities: outlier detection and embedding comparison. They enable the system to identify novel cell types and assess the similarity between different cell embeddings effectively.

Outlier detection identifies cells that significantly deviate from the majority of the dataset. This tool uses a distance-based approach to measure the dissimilarity between a target cell embedding and the centroid of existing cell type clusters. Specifically, it employs the Euclidean distance, which calculates the straight-line distance between two points in the feature space:

\begin{equation}
D_E(x, \mu) = \sqrt{\sum_{i=1}^{n} (x_i - \mu_i)^2}
\end{equation}

where \( x \) represents the target cell embedding, \( \mu \) denotes the mean vector of the cluster, and \( n \) is the dimensionality of the embedding space. A cell is flagged as an outlier if its Euclidean distance from the cluster centroid exceeds a predefined threshold. This threshold is dynamically adjusted based on the dataset's characteristics, ensuring robust detection across diverse biological contexts.The outlier detection tool is particularly useful for identifying potential novel cell types. By isolating cells that lie far from known clusters, scAgent can prompt further investigation into whether these cells represent previously uncharacterized cell types.

Embedding comparison quantifies the similarity between two cell embeddings, facilitating the identification of closely related cell types. The tool utilizes cosine similarity, a metric that measures the angle between two vectors in the feature space:

\begin{equation}
\text{similarity}(x, y) = \frac{x \cdot y}{\|x\| \|y\|}
\end{equation}

where \( x \) and \( y \) are the embeddings of two cells. A similarity score close to 1 indicates high similarity, while a score near 0 suggests significant dissimilarity. This metric is particularly effective in high-dimensional spaces, where it captures subtle differences in gene expression patterns. The embedding comparison tool integrates with the memory module to retrieve and compare embeddings from previously annotated cells. This integration enables scAgent to make informed decisions about novel cell types by referencing memory embeddings. 

\paragraph{Incremental Training Tools}

The incremental training tools integrated into scAgent facilitate the efficient extension to novel cell types. It leverages a limited number of cells from novel type to update the model parameters, thereby enabling the model to accurately annotate cells belonging to this type next time.

To achieve this functionality, we first extends the dimensionality of the classification head of scAgent to $C_{new} = C_{original} + \Delta C$, where $\Delta C$ denotes reserved capacity for novel cell types. Specifically, we set $C_{new}=200$ in implementation, substantially exceeding current cell category counts. Experimental validation confirms this dimensional expansion preserves prediction accuracy while enabling dynamic scalability as novel cell types emerge.


tDuring incremental training, we freeze base model parameters $W_0$ and exclusively optimize the MoE-LoRA module parameters $\Delta W_{MoE-LoRA}$. This strategy reduces trainable parameters  compared to full-model fine-tuning, significantly enhancing training efficiency. The frozen base parameters $W_0$ maintain existing knowledge representations, effectively mitigating catastrophic forgetting in conventional machine learning paradigms \cite{catastrophic}.

What's more, the tool implements a dynamic plugin update mechanism where newly acquired knowledge is encapsulated in MoE-LoRA components.

\subsubsection{The Memory Module of scAgent}

Complementing the tool hub, the memory module serves as a dynamic knowledge repository, enabling scAgent to retrieve and store critical information efficiently. This module comprises three core components: dataset, embedding and history (Fig.\ref{scagent_overview}d). 

The dataset component integrates both published reference datasets (e.g., CELLxGENE\cite{cellxgene}, Tabula Sapiens\cite{tabula_sapiens}) and dynamically uploaded user data. The stored data provides foundational support for MoE-LoRA plugins training. 

The embedding component is implemented using the Milvus vector database \cite{milvus}, an open-source platform optimized for managing and querying large-scale vector data. Milvus excels in scalability, low-latency performance, and high-dimensional similarity searches, handling billions of vectors with real-time responsiveness—ideal for scRNA-seq data analysis. It supports flexible deployment (standalone or distributed) and multiple indexing algorithms for diverse workloads.

In this system, Milvus organizes embedding vectors from scRNA models into two subspaces:  
(1) Standard embeddings: Pre-trained model outputs as foundational representations;  
(2) LoRA-enhanced embeddings: Task-specific representations from MoE-LoRA plugins for improved adaptability.
Milvus supports a variety of efficient indexing methods, and in this case, we utilize the IVF-FLAT indexing method to store vectors. IVF-FLAT is an inverted file index that partitions the vector space into clusters using a k-means algorithm,assigning each vector to the nearest cluster centroid. For querying, Milvus employs approximate nearest neighbor (ANN) search \cite{ANN}, calculating similarity (e.g. Euclidean distance) and using pre-built indices to quickly retrieve the top k nearest embeddings. This enables scAgent to perform real-time novel cell detection across large datasets with high efficiency.

The history component tracks query logs, tool execution traces, and cached intermediate states, providing contextual information for the planning module. 
Together, these components enable scAgent to maintain a dynamic and scalable knowledge base, supporting continuous learning and efficient analysis of scRNA-seq data.

\subsection{Novel Cell Detection}

During the process of nocel cell detection, scAgent leverages two feature extractors to get the embeddings. The first extractor, \( F_g \), is pre-trained on large-scale single-cell data using self-supervised learning (SSL) to capture a general-purpose representation of cellular diversity. The second, \( F_s \), is fine-tuned on labeled data of known cell types, initializing from \( F_g \), to produce embeddings optimized for specific classification tasks. For a gene expression matrix \( X \in \mathbb{R}^{n \times g} \) (where \( n \) is the number of cells and \( g \) is the number of genes), the embeddings are generated as:

\begin{equation} 
E_g = F_g(X) \in \mathbb{R}^{n \times d}, 
\end{equation} 
\begin{equation} 
E_s = F_s(X) \in \mathbb{R}^{n \times d}, 
\end{equation} 

where $ d $ represents the dimension of both the general and specific embedding spaces.

The tool utilizes two vector databases, \( D_g \) and \( D_s \), corresponding to \( F_g \) and \( F_s \), to store embeddings of all individual cells from known cell types, denoted \( C = \{C_1, C_2, \dots, C_k\} \) for \( k \) classes. For a cell \( x_{i,j} \) in class \( C_i \) (where \( j = 1, 2, \dots, |C_i| \) and \( |C_i| \) is the number of cells in \( C_i \)), the embeddings are:
\begin{equation} e_{g,i,j} = F_g(x_{i,j}), \end{equation} 
\begin{equation} e_{s,i,j} = F_s(x_{i,j}). \end{equation}

Thus, the databases are defined as \( D_g = \{e_{g,i,j} \mid i = 1, \dots, k; j = 1, \dots, |C_i|\} \) and \( D_s = \{e_{s,i,j} \mid i = 1, \dots, k; j = 1, \dots, |C_i|\} \).

For a novel cell with expression profile \( x_{\text{novel}} \), the tool computes its embeddings:
\begin{equation} e_{g,\text{novel}} = F_g(x_{\text{novel}}), \end{equation} \begin{equation} e_{s,\text{novel}} = F_s(x_{\text{novel}}). \end{equation} 

Nearest-neighbor retrieval is performed in \( D_g \) and \( D_s \) to identify the most similar known cells by calculating similarity distances in scRNA-seq data:

\begin{equation}
s_{g,i,j} = d(e_{g,\text{novel}}, e_{g,i,j}),
\end{equation}
\begin{equation}
s_{s,i,j} = d(e_{s,\text{novel}}, e_{s,i,j}),
\end{equation}

for all \( e_{g,i,j} \in D_g \) and \( e_{s,i,j} \in D_s \).The top-\( m \) nearest neighbors (e.g., \( m = 10 \)) are retrieved, forming sets \( N_g = \{(C_{g1}, s_{g1}), \dots, (C_{gm}, s_{gm})\} \) and \( N_s = \{(C_{s1}, s_{s1}), \dots, (C_{sm}, s_{sm})\} \), where \( C_{gj} \) and \( C_{sj} \) are the cell types of the nearest neighbors, and \( s_{gj} \) and \( s_{sj} \) are their similarity scores.

\begin{figure}[H]
\centering
\includegraphics[width=\textwidth]{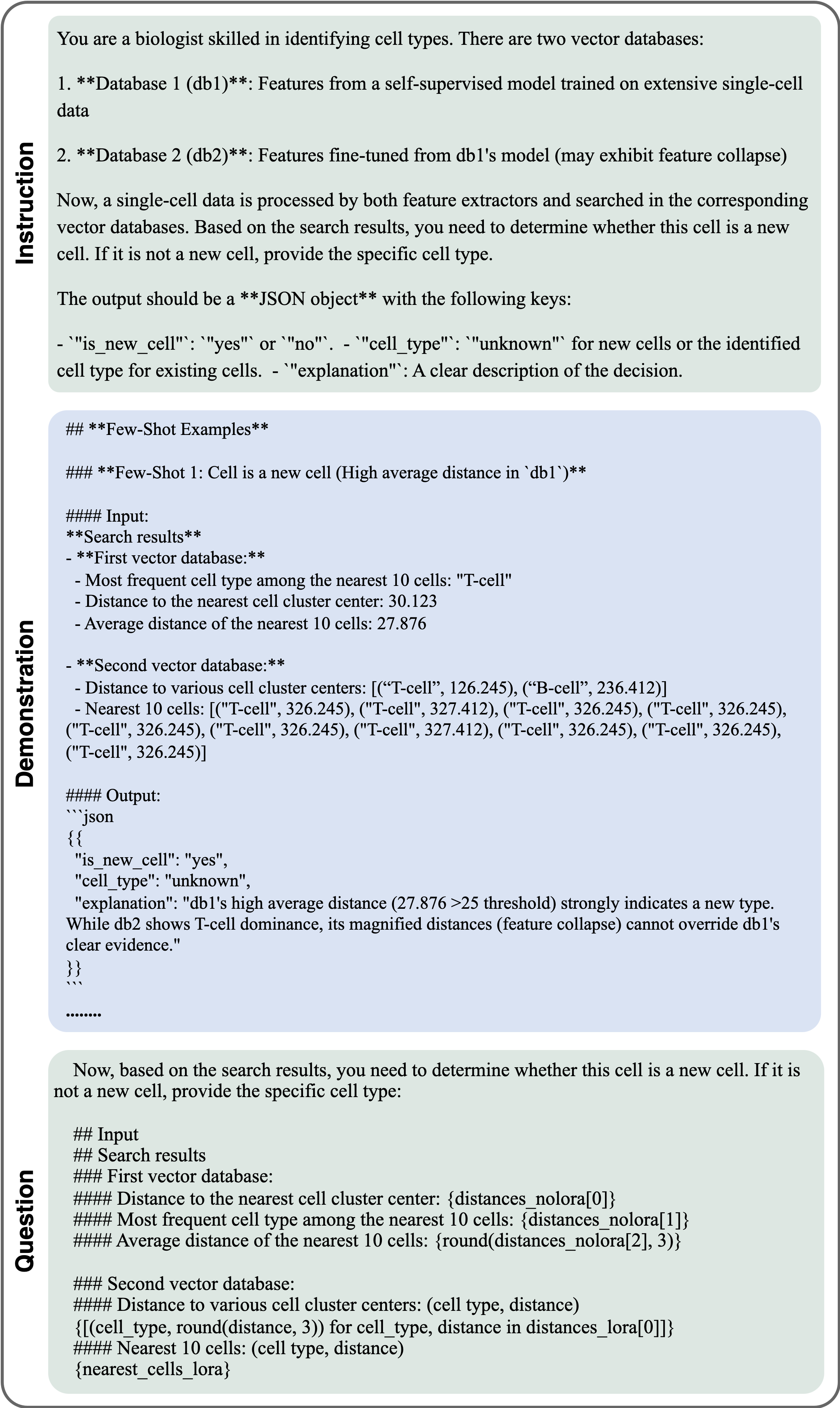}
\caption{Prompt Template for novel cell detection.}\label{prompt-openset}
\end{figure}

The retrieval results \( N_g \) and \( N_s \) are formatted as structured prompts and provided to an LLM to assess whether \( x_{\text{novel}} \) represents a novel cell type. The LLM integrates multi-source information, excelling when embeddings of novel and known cells are close. The prompt is a natural language description of the retrieval outcomes,  organized into three sections: instruction, demonstration, and question. The instruction clearly outlines the task of determining whether a single-cell sample is a novel cell type or an known one, using search results from two vector databases. The demonstration section provides four examples that illustrate how to analyze distances and cell type patterns to reach a decision. Lastly, the question section presents placeholder search results from both databases, prompting the LLM to deliver a JSON-formatted decision with an explanation based on the provided data. The prompt in Fig.\ref{prompt-openset} effectively guides the LLM to make an informed judgment and justify its reasoning.

\subsection{Data Preparation}

We prepare two primary datasets for our experiments: the CELLxGENE dataset (CG dataset) and the Tabula Sapiens dataset (TS dataset). Both datasets are widely used in scRNA-seq research and provide comprehensive coverage of human tissues and cell types.

The CG dataset is derived from CELLxGENE platform\cite{cellxgene}, following the download and preprocess outlined in \cite{sctab}. To ensure compatibility with our hardware setup (NVIDIA A40 46G, NVIDIA GeForce RTX 4090 24G, and 250 GB of memory), we perform downsampling on the raw dataset. Specifically, we apply stratified sampling based on cell types and manually adjust the sample sizes to ensure that each cell type and tissue is proportionally represented. This approach minimizes bias and ensures that the downsampled dataset retains the diversity of the original data.

The TS dataset is sourced from the Tabula Sapiens project\cite{tabula_sapiens}, which provides a detailed reference atlas of nearly 500,000 cells across 24 human tissues. This dataset offers finer-grained annotations compared to the CG dataset, enabling deeper insights into cell states and differentiation processes. We preprocess the TS dataset using the steps described in \cite{scgpt}, which include normalization, filtering, and quality control. These steps ensure that the dataset is suitable for training and evaluation.

We split both the CG and TS datasets into training, validation, and test sets using an 8:1:1 ratio. This split ensures that all cell types are represented in each subset, maintaining the integrity of the data distribution.

\subsection{Model Training}

For scAgent, we freeze the original parameters of the scRNA model and focus on training the MoE-LoRA plugins. Each plugin consists of 5 experts, with each expert corresponding to a LoRA module of rank 8. The MoE-LoRA plugins are divided into two categories: tissue-specific MoE-LoRA plugins and tissue-assignment MoE-LoRA plugins. The tissue-specific plugins are trained using data from specific tissues, with the training target being the cell types within each tissue. We set the classification head dimension to 200, exceeding the current 162 cell types, ensuring plug-and-play compatibility and enabling seamless integration of novel cell types during incremental training. For the tissue-assignment MoE-LoRA plugin, which assigns input cells to their respective tissues, we jointly train the MoE-LoRA components and the classification head to improve accuracy, with the training target being the tissue types. During training, we use the cross-entropy loss function to compute the loss, choose Adam as the optimizer. To stabilize training, we apply cosine learning rate decay, which prevents overly large steps in the later stages of training. 

For scGPT, we initialize the model with the pre-trained weights provided in \cite{scgpt}. We freeze the original parameters and only train the MoE-LoRA architecture and an MLP classification head, which we extend to 200 dimensions to accommodate the multi-classification task. The model is trained on the full dataset, with the training target being the cell types. Similarly, for scBERT, we extend its classification head to 200 dimensions to support the multi-classification task. For scTab (10X data), we use the weights provided in \cite{sctab}, which were trained on the raw CG dataset before downsampling. However, since its classification head weights are fixed, it cannot generalize to the TS dataset. For the standard scTab, we modify its classification head dimensions to fit the multi-classification task and retrain it on both datasets. For MLP, we set the number of hidden layers to 2 to prevent overfitting. All baseline methods follow their original training configurations and use their recommended default parameters. We adopt an early stopping strategy and train all models until convergence.

\bibliography{sn-bibliography}

\end{document}